# Entropy-based Optimization via A* Algorithm for Parking Space Recommendation

*Xin Wei[1], Runqi Qiu[1], Houyu Yu[2*], Yurun Yang[1], Haoyu Tian[1], Xiang Xiang[3*]*

{[1]Sch. of International Education, [2]Sch. of Automotive Eng.}, Wuhan University of Technology
[3]Sch. of Artificial Intelligence and Automation, Huazhong Univ. of Science and Technology
Wuhan, China
[1] 282428@whut.edu.cn, [2*] yuhouyu@sina.com, [3*] xex@hust.edu.cn

***Abstract*—This paper addresses the path planning problems for recommending parking spaces, given the difficulties of identifying the most optimal route to vacant parking spaces and the shortest time to leave the parking space. Our optimization approach is based on the entropy method and realized by the A* algorithm. Experiments have shown that the combination of A* and the entropy value induces the optimal parking solution with the shortest route while being robust to environmental factors.**

## I. INTRODUCTION

The key technology of the intelligent parking lot is the parking space recommendation [1]. Jun developed an APGM for guiding active parking guidance, which is still a framework with many limitations [2]. Shin studied a smart parking guidance algorithm with parking space's utility functions [3], but his algorithm cannot recalibrate the weighting function. This paper constructs an optimized parking space recommendation (OPSR) approach based on the A* algorithm and the entropy method to solve the problems above. Important factors include the vehicle cost, the distance from the entrance to the parking space, the distance from the parking space to the exit, and the reliability of the parking lot [4]. However, the OPSR approach in this paper is created for the management of a single parking lot, so there is no need to consider different rate of charge and the reliability of parking lot. Notably, the premises of a parking recommendation algorithm are based on normally include:

*Correspondence. This work is supported by Fundamental Research Funds for Chinese Central Universities, under Grant No: 2020-GJ-BI-IO

(1) The layout of parking spaces in the parking lot and the location of the road is known;

(2) The system can operate normally, and the system will not stop working due to changes in the external environment;

(3) The parking lot does not contain unreasonable parking situations, such as one car occupies two parking spaces.

In the following, related works are first introduced, and the second part illustrates the problem statement. Then the third section focuses on proposed approach. Next, an example analysis will be applied to illustrate the parking recommendation algorithm and compare the algorithm with existed approaches. Finally, this article summarizes and evaluates the algorithm. In summary, the contributions of this work include:

(1) Simplify the nodes to calculate by combining A* with undirected graph.
(2) Develop new index for evaluating the parking environment.
(3) Utilize fuzzy set to decline the effect of magnitude's order.
(4) Develop the self-regulating algorithm based on entropy method.

## II. RELATED WORKS

Since PGIS was first proposed by German city of Aachen[6], researchers have conducted immense research in the field of intelligent parking lots and obtained various valuable results[1]. Wu *et al.* compared three different existing intelligent parking systems and summarized their

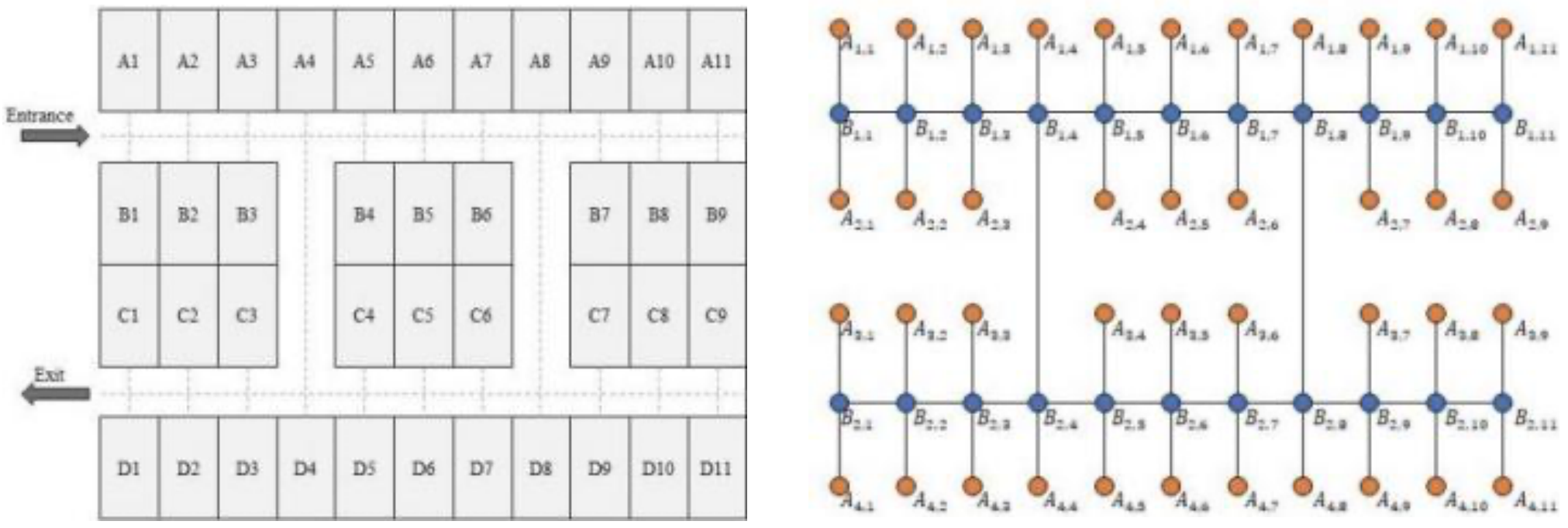

(a) (b)

Fig.1 parking map

advantages and disadvantages[4], including the RFID-based system, the system based on ultrasonic sensing technology, and the system based on electromagnetic induction coil[4]. They demonstrated a suitable method to construct the intelligent parking system at the hard-ware level[4]. Meanwhile, Han *et al.* clarified the realization of the smart parking lot at the software level[5]. Specifically, theydivided the WiFi-based parking management system into three modules, including parking lot management, user management, as well as the space allocation and route arrangement, which proved the feasible implementation of the system[5]. On the basis of achieving both the hardware level and the software level, Fu et al. proposed multiple evaluation indicators to control the parking space recommendation system[6]. Consequently, the system recommended users with the parking set, which is able to better satisfy the customers' needs[6][7]. Meanwhile, Leephakpreeda et al. proposed to use fuzzy knowledge decision-making to control the parking guidance system[8], which allows the evaluation index proposed by Fu et al.[6] to be more widely applied. With technical support, the intelligent parking lot has the ability to recommend basic parking spaces [8]. Despite the research on single parking lots, the research on smart parking has been extended to further assess the methods of smart parking management[1][2]. In particular, it will analyze methods of allowing multiple cars to park in the different parking spaces in the shortest possible time through the parking prediction model [2].

## III. PROBLEM STATEMENT

The difficulties of recommending parking space faces mainly lie three aspects:

(1) The factors are not sophisticated, they cannot evaluate the parking situation for reversing and parking the cars.

(2) The searching algorithm is not mature, which can be simplified to be efficient.

(3) The weighting function cannot be recalibrated to adapt to environment.

Algorithm still needs to promise to classic indexes, including the optimal driving duration, driving distance and walking distance, etc. The method to obtain these indexes is by analyzing the parking map Fig.1(a), and it is always simplified by points and lines in Fig.1(b). A, B, C, and D represent four parking rows, respectively. $B_{i,j}$ represents the intersection, $A_{i,j}$ represents the parking space.

## IV.PROPOSED METHOD

The OPSR approach needs to be based on relevant influencing factors. Thus, this paper introduces a weight function to simultaneously introduce multiple factors.

$$H = W_1 \times X + W_2 \times L + W_3 \times S \tag{1}$$

The weight function (1) refers to multiplying different influencing factors X, L and S with corresponding weights W1, W2 and W3 to obtain a comprehensive index, so that the environment and location of different parking spaces are evaluated. The following section specifically describes each factor that affects the parking space recommendation system.

(1) Vehicle driving distance X: the distance from the entrance of the parking lot to the target parking space, the unit is m.

(2) Pedestrians walk out of distance L: the distance where the pedestrian walks from and the target parking space to the exit of the parking lot, the unit is m.

(3) Parking situation of surrounding vehicles S: the parking situation of vehicles on both sides of the parking space does

not have a unit. The specific evaluation method for this factor focuses on classifying and discussing the possible situations, as shown in Fig. 2. When there are vehicles on both sides (a), only one side (b), and on both sides (c), the difficulty of parking is set to 3, 2, 1, respectively.

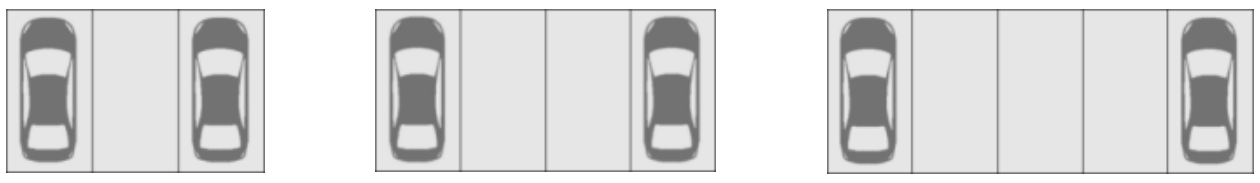

(a) (b) (c)

Fig. 2. Situation of different parking sets.

Finally, the parking space with the best comprehensive index will be regarded as the optimal recommendation of parking spaces to users. The influencing factor indices are standardized with fuzzy sets, and the weight values are then obtained by the entropy method, and the comprehensive index can finally be obtained.

### *A. A* Algorithm Calculates the Shortest Path*

For the calculation of the first factor—vehicle driving distance and the second factor—pedestrian walking distance, this paper utilizes the A* algorithm to solve it. A* algorithm has developed many years and it has various optimal method to improve the algorithm[10]. For smart parking lots, the map can be processed as an undirected graph. Take the Fig.1(b) as an example. First, we put the parking map into a coordinate system for locating in Fig.3.

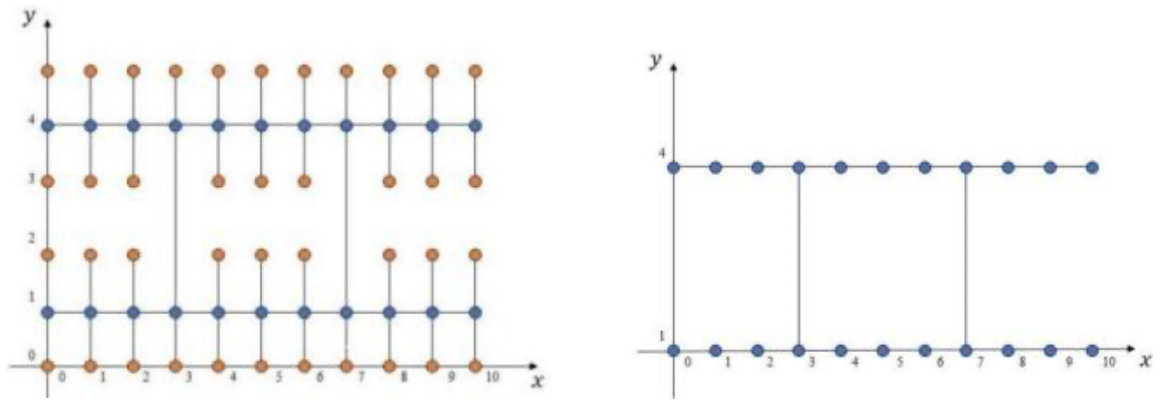

(a) (b)

Figure 3.Parking coordinate system

First, we put the parking map into a coordinate system for locating in Fig.3. To simplify the problem, the intersection and the parking space can be regarded as the same nodes. In the simplified undirected graph, the reduction of the number of nodes greatly reduces the time it takes to traverse all nodes greatly reduced. For the calculation of X, it is necessary for algorithms to add the target parking space to the OPEN list, and then diverge around to find the node with the least cost of function f(n). When the node of divergent expansion is the entrance of the parking lot, the loop ends, and then backtracks to the parent node until the target parking space is backtracked. After obtaining the path generated in the backtracking process, the A* algorithm calculates the length of the entire path through the length attribute of the passed edge. The calculation process of L is similar to that of X.

### *B. Standardization of Influencing Factors*

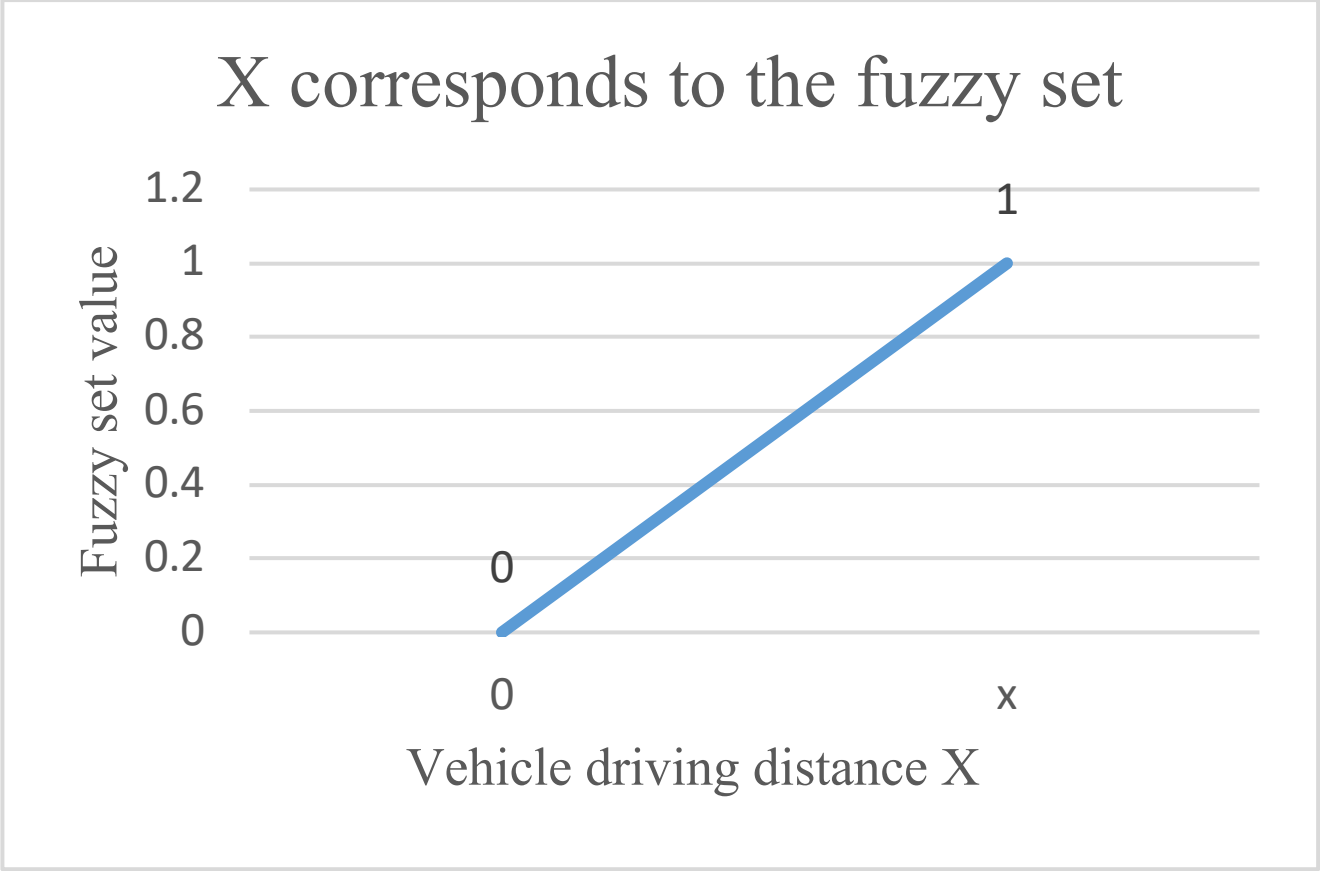

Fig. 4 Fuzzy set curve corresponding to vehicle travel distance **X**

After obtaining the three relevant factors, the data is standardized by means of fuzzy sets. The variable model set is described in details [11]. Fuzzy set U refers to a mapping U , from U to unit interval [0,1] given a universe of discourse $\mu_A\colon\ U \to [0,1]$ . Consequently, the first related factors mentioned above can be numerically standardized, as shown in Fig. 4. The x on the abscissa axis in Fig. 4 refers to the distance of the farthest vehicle (take the entrance of the parking lot as the starting point and the parking space farthest from the entrance of the parking lot as the end point). Through the fuzzy set, we treated the vehicle driving distance X, the pedestrian walking distance L, and the surrounding parking situation S as a set with the highest value of 1.

### *C. Entropy Method to Determine the Weights*

The specific method of using the entropy method to calculate the weight is to first initialize the data(2).After obtaining each Yi,j value, a standardized matrix can be formed (3).

$$Y_{ij} = \frac{X_{ij}}{\sum_{i=1}^{m} x_{ij}}, \quad 0 \le Y_{ij} \le 1 \tag{2}$$

$$Y = \{y_{ij}\}_{m \times n} \tag{3}$$

Since the unit entropy function is in this form(4), the unit entropy value function can be used to obtain the information entropy value of j items(5).

$$e = -k \sum_{i=1}^{m} y_i \ln y_i \tag{4}$$

$$e_j = -k \sum_{i=1}^{m} y_{ij} \ln y_{ij} \tag{5}$$

To get ej, k needs to be determined. The constant k in (5) is related to the number of samples m of the system. This is because when the information of the system is completely disordered, the system entropy value is 1, and in this completely disordered situation, when the number of samples is m, $y_{ij} = \frac{1}{m}$. Then, through combining these two known quantities, it can be obtained by (6):

$$k = (\ln m)^{-1}, 0 \le e \le 1 \tag{6}$$

$$h_j = 1 - e_j \tag{7}$$

To obtain the amount of information value used to judge a certain index, we need to use (7). The larger the information utility value coefficient hj, the more significant the indicator is for judgment, so the weight can be determined by (8):

$$w_j = \frac{h_j}{\sum_{j=1}^{n} h_j} \tag{8}$$

Based on the above method, the specific operation method of the system is depicted in the following Fig.5:

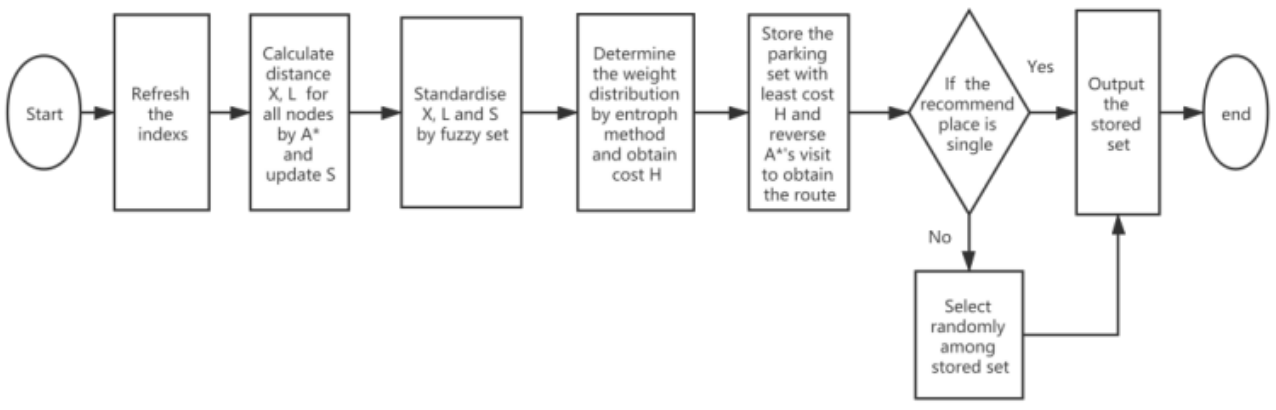

Fig.5. Algorithm flow chart

## V. EXPERIMENTS

To better demonstrate the parking space recommendation system proposed in this paper, an example which features a parking lot plan will be discussed.

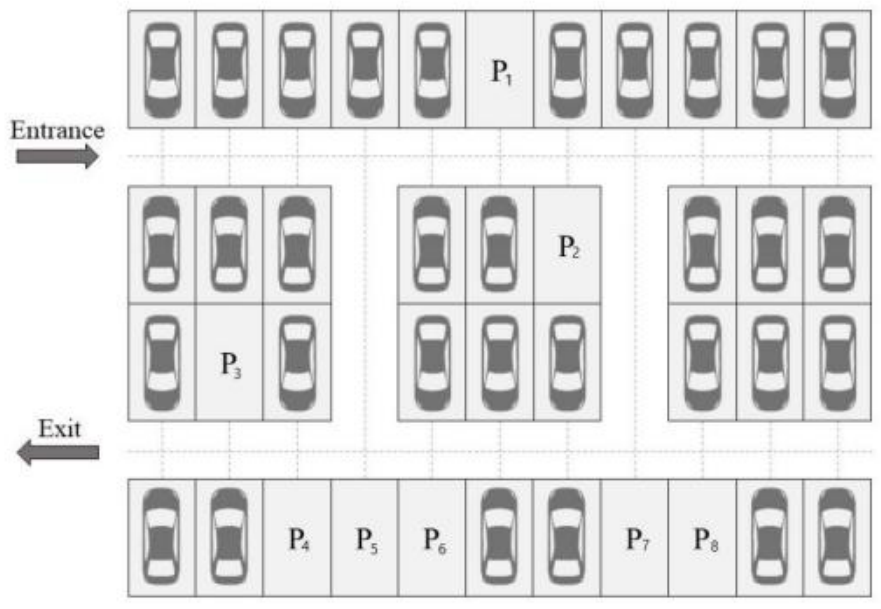

Fig. 6. Layout plan of a parking lot

In this parking lot plan, the entrances, exits, roads, parking spaces and available parking spaces of the parking lot are assigned. Each standard parking space is 2.4m wide, 5.3m long, and the road width measured to be 6m. The specific location of the vacant parking space is depicted in Fig.6, including P1, P2, P3, P4, P5, P6, P7, P8. The relevant attributes of these eight parking spaces are represented by the influencing factors mentioned above as shown in Fig. 7.

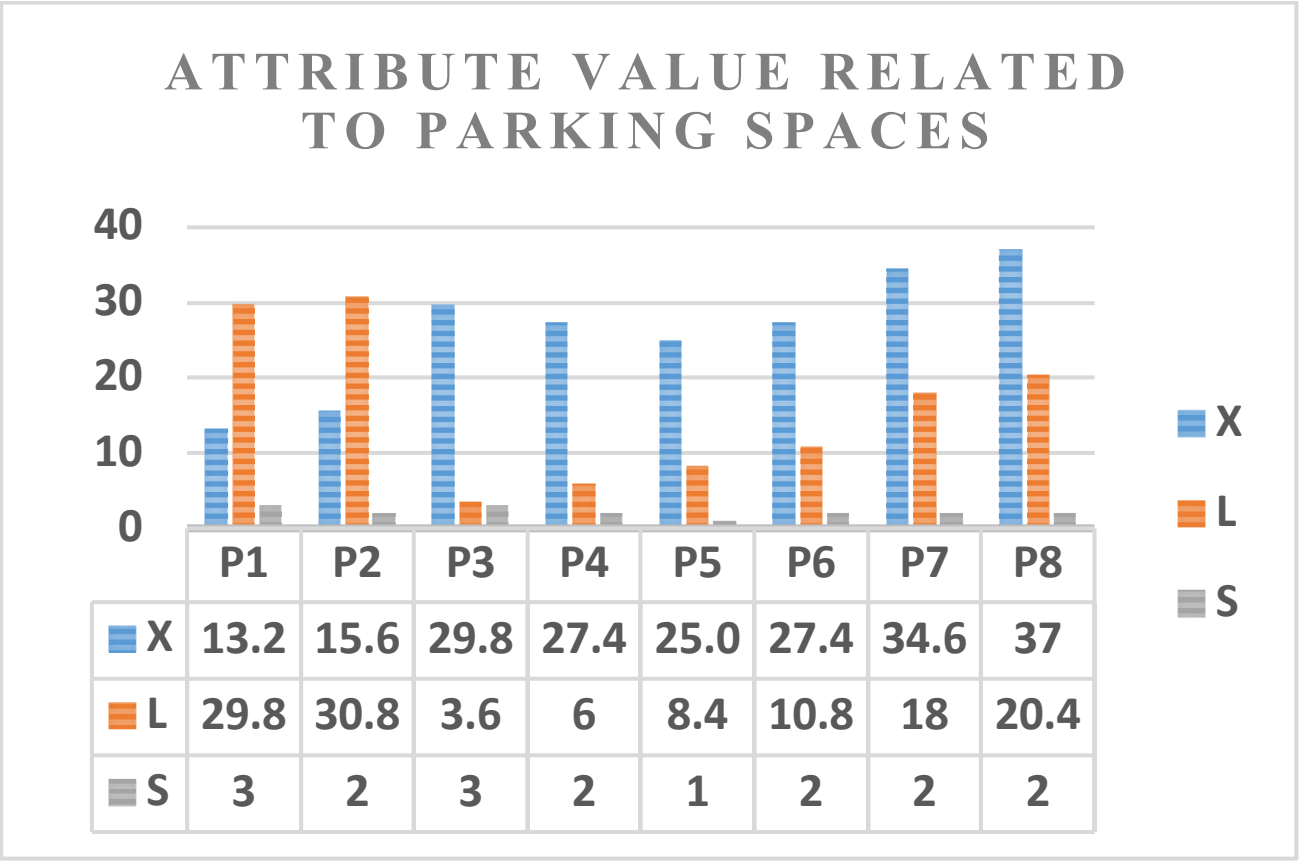

Fig. 7. Related attributes of parking spaces

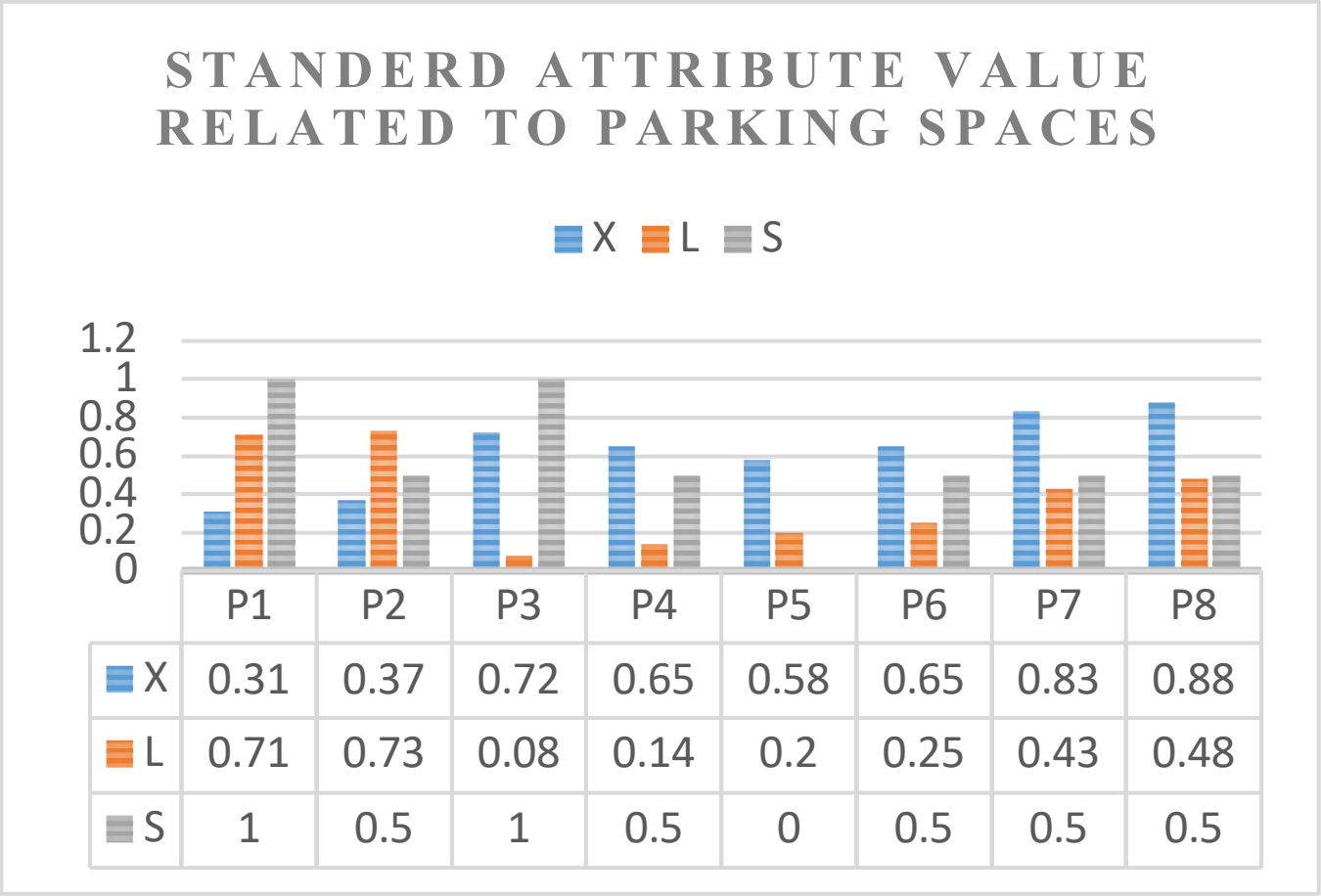

Fig. 8. Standardized values of relevant attributes of parking spaces

After obtaining the value of the original attribute of each parking space, the value needs to be standardized, then the fuzzy set method needs to be applied. The distance from the parking lot entrance to the farthest parking space is 41.8m, and the distance from the parking lot exit to the farthest parking space from the exit is 41.8m. The normalized value is depicted in Fig. 8.

The large amount of data obtained at this time will act as the basis for the entropy method in determining the weight. First, the logarithmic value needs to be further standardized according to (2) and (3), and it is then formed into a matrix (9).

(9)

After obtaining the constant K by formulas (5) and (6), the information entropy e of each indicator can be calculated, and the utility value h of the information can be obtained. Under such circumstances, we get $e_1 = 0.97$, e2 = 0.906, $e_3 = 0.972$, $h_1 = 0.025$, $h_2 = 0.094$, $h_3 = 0.028$. Here, the formula (8) can be used to calculate the weight W of each evaluation index.

which are $W_1 = 0.17$, $W_2 = 0.64$, $W_3 = 0.19$.

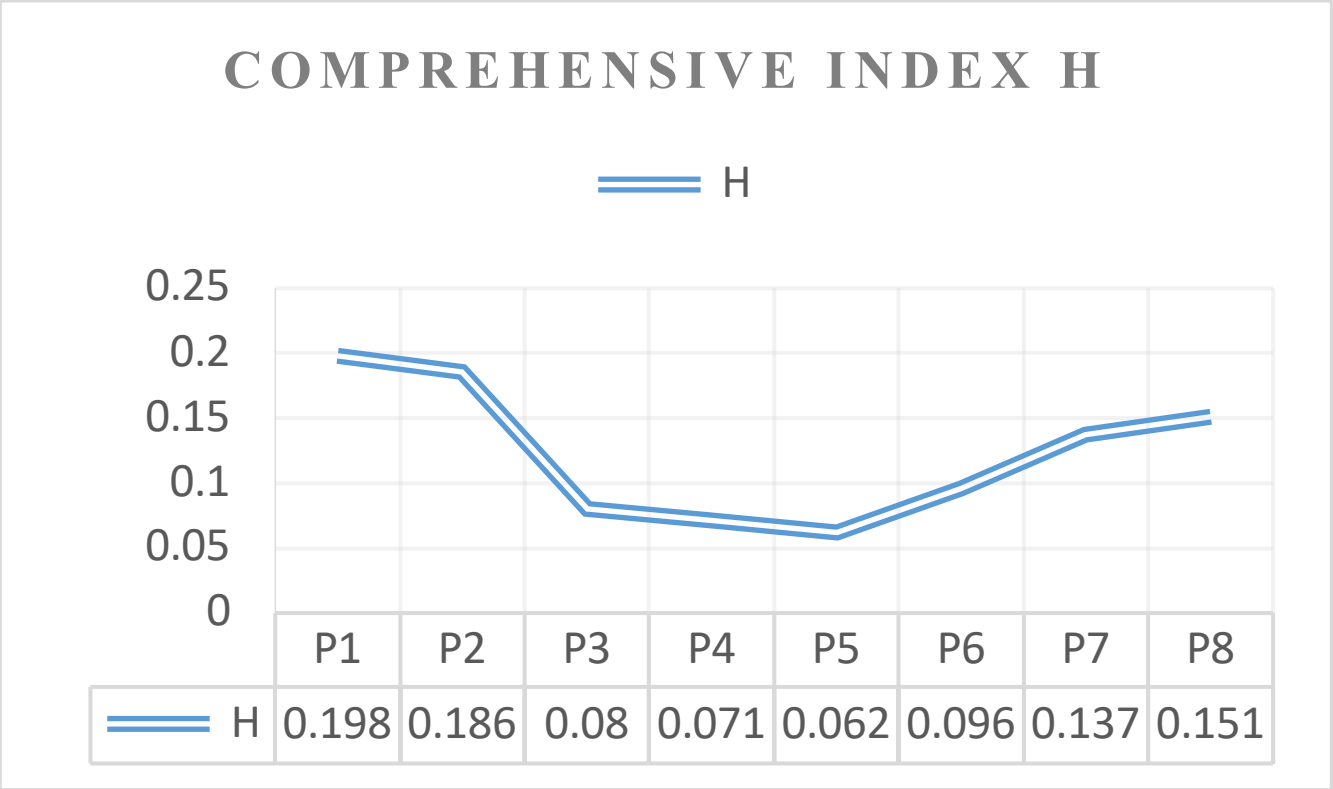

Fig. 9 Comprehensive index value of each available parking space

Finally, formula (1) can be used to obtain the comprehensive index of these eight parking spaces, as illustrated in Fig. 9. Since the comprehensive index H of parking space P5 is the smallest, the recommended parking space of the system will be P5.

The experiment compares OPSR approach with the recommendation algorithm mentioned in the YAN's paper[5] and four weighting algorithms with different tendencies: I. W1 = 1, W2 = 1, W3 = 1; II. W1 = 10, W2 = 1, W3 = 1; III. W1 =1, W2 = 10, W3 = 1; IV. W1 = 1, W2 = 1, W3 = 10 in four typical different environments to prove the advantages of the OPSR approach: A. Vacant B. Every place is surrounded by two cars C. Control the same distance for driving(only C3, C4, C5, D3, D5 are vacant parking spaces) D. Control the same distance for walking (only A3, A5, B3, B4, and B5 are empty parking spaces). Shin assessed the approach by various measures[10]. This paper utilizes the critical index Duration (s). In the model, the speed of pedestrians is set to 1.1m/s, the speed of the car is 5km/h, and the corresponding time for parking difficulty from high to low is 210s, 157.5s, 105s.

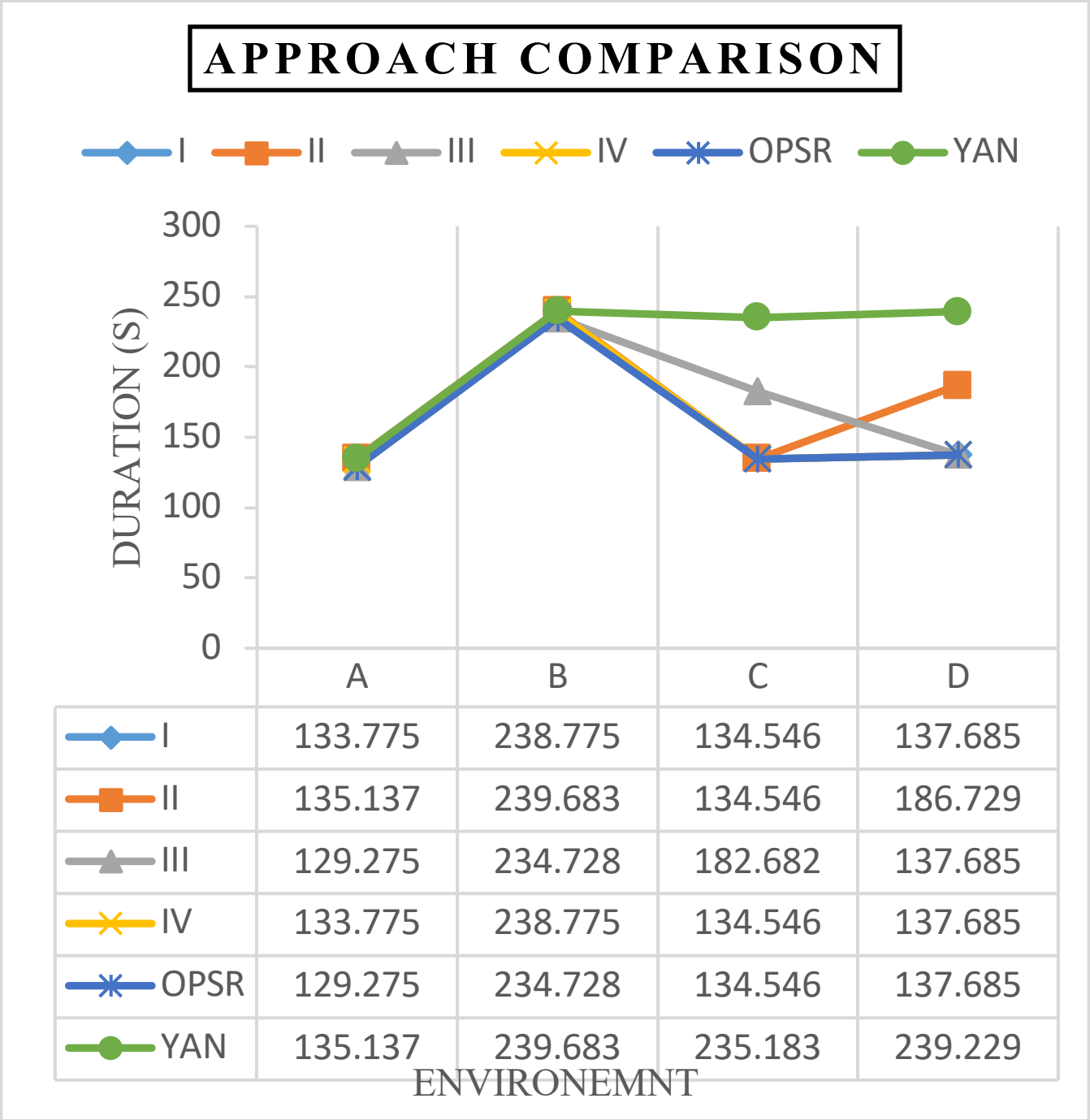

Fig. 10 Approach comparison

The results show the importance of adjusting the weight equation in different environments by Fig. 10. The total time consumption of the OPSR solution is always the lowest in different typical environments, when it is compared with the existing YAN's solution[5] and other solutions.

## VI. CONCLUSION

This article introduces an algorithm for implementing smart parking space recommendation. After exploring the relevant factors that affect the user experience, the parking recommendation algorithm in this paper uses the combination of the A* algorithm, undirected graph and situation analysis to obtain the value of each influencing factor. Then, it applies the concept of fuzzy set to standardize the data and utilizes the entropy method. The reasonable weight distribution method is obtained through these methods, and finally a comprehensive index is produced through the proposed weight function, and parking spaces based on the comprehensive index as an indicator are recommended. The algorithm reduces the computation by simplifying the nodes with undirected graph based on A* and standardizing the value by fuzzy set. The algorithm can adjust the weight of factors to fit the changing environment by entropy method. However, the entropy value requires adequate data for analysis, and the A* algorithm is perhaps not able to support the solution in specific situation.